\documentclass[11pt,a4paper,english,11pt,a4paper]{article}
\usepackage[utf8x]{inputenc}
\usepackage{fancyhdr}
\usepackage{color}
\usepackage{babel}
\usepackage{graphicx}
\usepackage[authoryear]{natbib}
\usepackage{subscript}
\usepackage[unicode=true,pdfusetitle,
 bookmarks=true,bookmarksnumbered=false,bookmarksopen=false,
 breaklinks=false,pdfborder={0 0 1},backref=false,colorlinks=true]
 {hyperref}

\makeatletter

\pdfpageheight\paperheight
\pdfpagewidth\paperwidth

\@ifundefined{date}{}{\date{}}
\usepackage[T1]{fontenc}
\usepackage{babel}
\usepackage{arabtex}
\usepackage{ucs}

\usepackage[hyperref]{acl2019}
\usepackage{times}
\usepackage{latexsym}
\usepackage[super]{nth}

\usepackage{url}
\usepackage{amssymb}
\usepackage{amsfonts}

\aclfinalcopy % Uncomment this line for the final submission
\title{Adversarial Generation and Encoding of Nested Texts}
\author{Alon Rozental \\
  Amobee Inc., Tel Aviv, Israel \\ \\
   \tt alon.rozental@amobee.com }

\makeatother

\begin{document}
\maketitle 
\begin{abstract}
In this paper we propose a new language model called AGENT, which
stands for \textbf{A}dversarial \textbf{G}eneration and \textbf{E}ncoding
of \textbf{N}ested \textbf{T}exts. AGENT is designed for encoding,
generating and refining documents that consist of a long and coherent
text, such as an entire book, provided they are hierarchically annotated
(nested). i.e. divided into sentences, paragraphs and chapters. The
core idea of our system is learning vector representations for each
level of the text hierarchy (sentences, paragraphs, etc...), and train
each such representation to perform 3 tasks: The task of reconstructing
the sequence of vectors from a lower level that was used to create
the representation, and generalized versions of the Masked Language
Modeling (MLM) and ``Next Sentence Prediction'' tasks from BERT
\citet{DBLP:journals/corr/abs-1810-04805}. Additionally we present
a new adversarial model for long text generation and suggest a way
to improve the coherence of the generated text by traversing its vector
representation tree.
\end{abstract}

\section{Introduction\label{sec:Introduction}}

Transformer \citet{DBLP:journals/corr/VaswaniSPUJGKP17} based neural
architectures have recently achieved great advancement in representing
and generating sequences of text \citeauthor{2019arXiv190102860D,noauthororeditor,Cer:2018aa,DBLP:journals/corr/abs-1810-04805}.
However, the tasks of representing and generating text that is both
long and coherent still eludes state of the art (SOTA) models. Specifically,
the effectiveness of these models decreases sharply when modeling
sequences that are longer than the span their Self-Attention layers
work upon. This problem was quantified in \citet{2019arXiv190102860D},
where it was referred to as \textit{Relative Effective Context Length}
(RECL), and it was shown that for relatively short self-attention
spans (128 tokens), the RECL can extend for up to 4.5 times the self-attention
span. 

A recent attempt to generate relatively long texts was presented in
\citet{noauthororeditor}, where a huge Transformer based language
model (1.5 billion parameters) was trained. This model was able to
successfully generate long texts of \textasciitilde 200 words that
seem nearly indistinguishable from human generated ones. However,
due to the computational complexity of the Transformer's Self-Attention
layer - $O\left(n^{2}\right)$ where n is the number of words - it
is very expensive to extend the length of generated text while maintaining
its coherence.

There have also been several recent attempts to encode long texts
as vectors, mainly for the purpose of using those vectors for text
classification tasks. In two such cases \citet{DBLP:journals/corr/abs-1810-04805,Cer:2018aa},
SOTA results have been achieved for multiple tasks, while using text
classification vectors of equal size to the (single token) embedding.
While both the \textless CLS\textgreater{} vector presented in \citet{DBLP:journals/corr/abs-1810-04805}
and the summation of the contextual token vectors in the document
presented in \citet{Cer:2018aa} have achieved impressive results,
the constraint of represent both words and long sequences of text
as vectors in the same space becomes harder as our text grows longer. 

\begin{figure*}[t]
\includegraphics[scale=0.22]{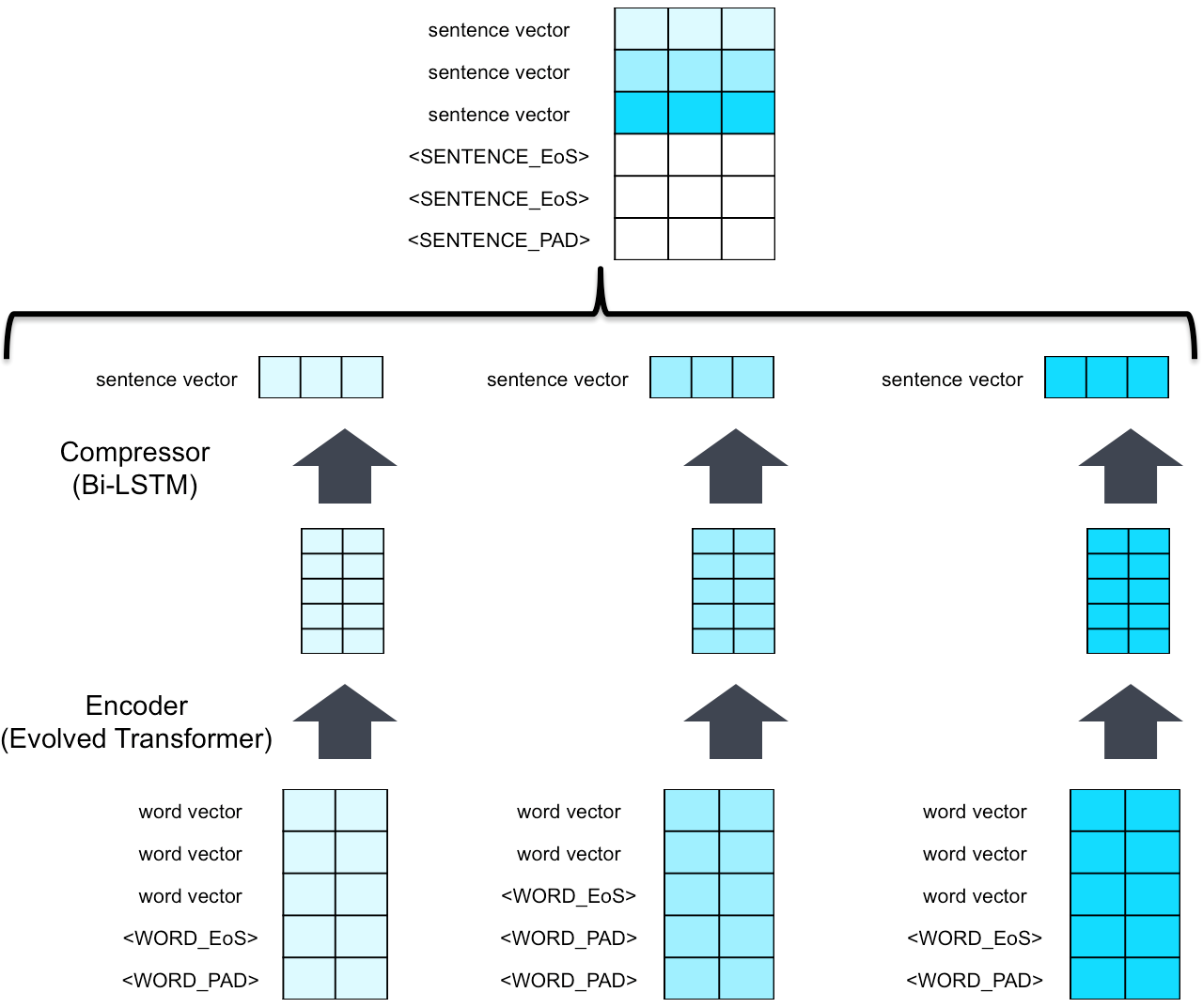}\caption{\label{fig:Hierarchal-Encoder-architecture}Hierarchal Encoder architecture.
EoS stands for ``end of sequence'' }
\end{figure*}

This paper is organized as follows: Section \ref{sec:Text-embedding}
describes how long texts are embedded. Section \ref{sec:Model-Training}
describes the training process required to make such embedding meaningful.
Section \ref{sec:Text-Generation} describes the generation and refining
processes for novel texts. Section \ref{sec:Proper-Nouns-Data} describes
who we keep proper nouns (such as character names) consistent throughout
the document, and Section \ref{sec:Summary} reviews the entire system
and its training process.

\section{Text embedding \label{sec:Text-embedding}}

In this section we explain how text is embedded into a one dimensional
vector given a pre-trained model. For this purpose, we assume that
the text was preprocessed, and that sentence, paragraph and chapter
breaks were identified during this stage. In order to encode a document
we only need to consider a small part of the network described here,
which we denote as the Hierarchal Encoder (HE). The HE creates a document
encoding according to the following steps: 
\begin{enumerate}
\item Tokens are embedded into $\mathbb{R}^{D_{token}}$, in a manner similar
to \citet{DBLP:journals/corr/VaswaniSPUJGKP17} using an embedding
matrix.
\item These tokens are transformed using a Self-Attention based Encoder.
Specifically, the Evolved Transformer Encoder model \citet{so2019evolved}
is used for this stage.
\item The encoded tokens are compressed into a 1 dimensional vector by a
Recurrent Neural Network (RNN), similar to \citet{Sutskever:2014:SSL:2969033.2969173},
with an output dimension of $D_{sentence}$, such that $D_{sentence}>D_{token}$.
For each sentence, the final state of this RNN is considered to be
the sentence representation vector, and we will refer to this RNN
as the Compressor. Please note that what we call a Compressor is referred
to as an Encoder in \citet{Sutskever:2014:SSL:2969033.2969173}. For
the RNN we use a Bidirectional LSTM inspired by \citet{article}.
\item All of the sentence vectors belonging to the same paragraph are stacked
and followed by a \textless SENTENCE\_EoS\textgreater{} token$\in\mathbb{R}^{D_{sentence}}$to
get a sentence embedding matrix.
\item The sentence embedding matrix is padded with trainable \textless SENTENCE\_PAD\textgreater{}
tokens, ensuring that all sentence embedding matrices are of the same
dimensionality.
\item A Position Embedding matrix is added to the sentence embedding matrix
in a manner similar to \citet{DBLP:journals/corr/VaswaniSPUJGKP17}.
\item A Segment Embedding matrix is added to the sentence embedding matrix.
During text embedding (though not necessarily during training) this
matrix is composed of a stack of identical (trainable) vectors, similar
to \citet{DBLP:journals/corr/abs-1810-04805}. See \ref{subsec:Coherence}
for further details.
\item Using the sentence embedding matrix, we repeat stages 2-6 to create
a paragraph encodings token $\in\mathbb{R}^{D_{paragraph}}$, such
that $D_{paragraph}>D_{sentence}$. 
\item Chapter embeddings are created by repeating the above process with
the paragraph as the input, and so forth.
\end{enumerate}
We denote all the intermediate vectors created by the HE during the
embedding of a document as the Document Vector Tree (DVT). The DVT
will help us during loss calculations in \ref{sec:Model-Training},
and figure \ref{fig:Hierarchal-Encoder-architecture} illustrates
the creating of a single (sentence level) node in that tree. 

\section{Model Training and Loss Function\label{sec:Model-Training}}

\begin{figure*}[t]
\includegraphics[scale=0.3]{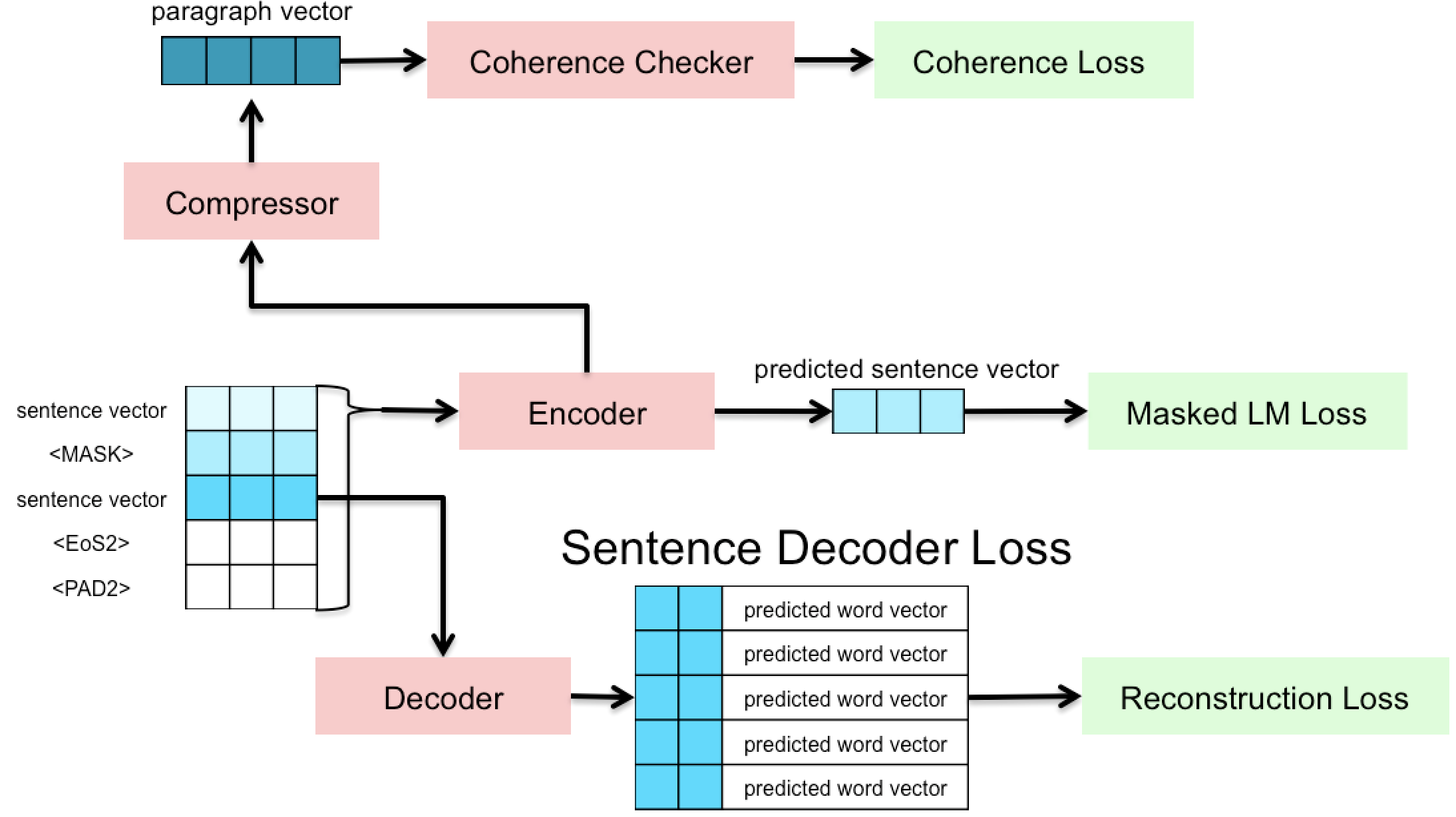}\caption{\label{fig:The-Three-Losses}The Three main losses of AGENT at the
sentence level.}
\end{figure*}

The loss function used to train our model contains three separate
components for each level of the DVT (except the first and last ones
that only have two). These components are referred to as:
\begin{itemize}
\item The Sequence Reconstruction (downward) loss
\item The MLM (in-level) loss
\item The Coherence (upward) loss
\end{itemize}
These losses are defined in the following subsections and their summation
is the main component in the loss function of AGENT. For these steps,
we will need to use the DVT. Thus, calculating it for each training
example (given the current weights of the HE) is the first step in
calculating these losses. Figure \ref{fig:The-Three-Losses} illustrates
these losses at the sentence level.

\subsection{Sequence Reconstruction\label{subsec:Sequence-Reconstruction}}

When we use a Compressor to create a higher level embedding from a
lower level one, for example creating a paragraph embedding vector
from its sentence embedding matrix, we would like to ensure that as
little data as possible is lost during the compression stage. To do
so, we use another RNN (also a Bidirectional LSTM) denoted as the
Decompressor, followed by a Self-Attention based Decoder that takes
the hidden states of the Decompressor as the input. Specifically,
we use the Evolved Transformer Decoder model \citet{so2019evolved}.
When reconstructing token vectors from sentence vectors, the loss
is computed in a way similar to \citet{DBLP:journals/corr/VaswaniSPUJGKP17}.
When reconstructing vectors at the sentence level (or higher), from
the output of their corresponding decoder, we make the following adjustments
in order to compute the loss in the same manner:
\begin{itemize}
\item The vector for the correct ``label'' of a sentence is the sentence
vector that was constructed by the HE.
\item The vectors for the incorrect ``labels'' (corresponding to the embedding
matrix in the token decoding stage) are all the sentence vectors created
by the HE for the current batch.
\item The bias for each sentence is remove, which is equivalent to setting
the bias vector of the token level decoder to 0. 
\end{itemize}

\subsection{Masked Language Modeling\label{subsec:Masked-Language-Modeling}}

For each level of the DVT, we perform the Masked Language Modeling
task described in \citet{DBLP:journals/corr/abs-1810-04805}. In this
task vectors (of tokens, sentences, paragraphs, etc...) are randomly
masked, and the objective is to predict the corresponding vector that
was generated by the HE. For the token level, we perform the task
using the exact same loss function as \citet{DBLP:journals/corr/abs-1810-04805}
and reduce mismatch between training and embedding by: 
\begin{itemize}
\item Replacing the missing vector with a vector corresponding to a \textless MASK\textgreater{}
token that exists in our vocabulary 80\% of the time.
\item Replacing the missing vector with a vector corresponding to a random
token that exists in our vocabulary 10\% of the time.
\item Keeping the ``missing'' vector unchanged 10\% of the time.
\end{itemize}
For higher levels of the DVT, we make the following adjustments:
\begin{itemize}
\item Replacing the missing vector with a \textless LEVEL\_MASK\textgreater{}
vector that is a learnable model variable 80\% of the time. Each level
of the DVT has its own ``mask'' token of the appropriate dimension.
\item Replacing the missing vector with a random vector, uniformly selected
from all of the sentence vectors generated by the HE for the current
batch 10\% of the time.
\item Keeping the ``missing'' vector unchanged, as it was generated by
the HE, 10\% of the time.
\end{itemize}
The classifier for this task shares some of its architecture and weights
with parts of the HE. For the sentence level the process is as follows:
\begin{enumerate}
\item Tokens are encoded and compressed by the token level Encoder and Compressor
of the HE to create sentence vectors.
\item A few of the sentence vectors are masked for the MLM task.
\item The sentence level Encoder of the HE is used to create contextual
embeddings. 
\item Like in \citet{DBLP:journals/corr/abs-1810-04805}, these contextual
embeddings pass through one dense layer with ``GELU'' \citep{2016arXiv160608415H}
activation (which is not a part of the HE) and go through a matrix
multiplication with the embedding matrix (which was generated by the
HE like in \ref{subsec:Sequence-Reconstruction}) to obtain the logits
for the classification task. 
\end{enumerate}

\subsection{Coherence \label{subsec:Coherence}}

This task is a generalization of what is referred to in \citet{DBLP:journals/corr/abs-1810-04805}
as the ``Next Sentence Prediction'' task and is meant to replace
it, both for the token level and higher levels of the DVT. In this
task (at the sentence level) all sentence vectors of the same paragraph
are randomly split into two groups: group A and group B. When creating
Segment Embeddings for the DVT (see \ref{sec:Text-embedding}), all
sentences are considered to be in group A. However, when performing
the Coherence task, a number P is randomly chosen. Half the time P
is set to 0, and half the time P\textasciitilde U(0,1). Then, each
sentence has a probability of P to belong to segment B. \textless SENTENCE\_PAD\textgreater{}
and \textless END\_OF\_PARAGRAPH\textgreater{} tokens are not considered
in this split and are always assigned to group A. Sentence vectors
belonging to segment B have a 50\% chance of being replaced by a random
sentence vector that was created by the HE in the current batch when
calculating the Coherence loss.

After this selection, we add a Segment Embedding to each sentence
vector, which is one of two special trainable tokens: \textless SENTENCE\_A\textgreater{}
and \textless SENTENCE\_B\textgreater , thus creating a new sentence
embedding matrix. We then create a paragraph vector using the sentence
level Encoder and Compressor of the HE. Note that in cases where P=0,
this paragraph vector is identical to the paragraph vector created
by the HE. This paragraph vector is passed to a component of our system
that we denoted as the Coherence Checker, which exists for each level
of the DVT. The Coherence Checker of each level ($CC_{level}$), is
a regressor that predicts the ratio of sentences in the paragraph
that were, in-fact, replaced. It is composed of L (a hyper-parameter,
see \ref{subsec:Generator}for more details) fully connected feed-forward
layers with tanh activation, followed by single unit with a sigmoid
activation. The $CC_{paragraph}$ loss is the mean-squared-error (MSE)
between the predicted and actual ratio of replaced sentences.

\subsection{Auto Encoder Regularization:}

In order to create one-dimensional text embedding vectors for the
various levels of the DVT we use components of our system that we
have called Compressors and Decompressors. While these components
perform a similar function to Encoders and Decoders, and in fact were
used as such in \citet{Sutskever:2014:SSL:2969033.2969173}, we treat
them as separate components from the Encoder and Decoder \citet{so2019evolved}
that we use along side them. The reason for that is that unlike the
Transformer Encoder, the Compressor must create a one-dimensional
output, while the Decompressor is not allowed to use any input but
the compressed vector such as the pre-compressed embedding.

While reversing the transformation of a Compressor is not necessary
to have an effective (Sequence Reconstruction loss minimizing) Decompressor,
we can safely say that a Decompressor which reverses its corresponding
Compressor is good enough, as it leaves us with the well tested Encoder
and Decoder of \citet{so2019evolved}. To encourage this effect, we
treat each corresponding Compressor-Decompressor pair as a part of
an Auto-Encoder and add the following regularization term to our model
for each such pair: 

\[
\epsilon_{auto}\cdot\parallel\frac{C_{i}}{\parallel C_{i}\parallel}-\frac{D_{o}}{\parallel D_{o}\parallel}\parallel
\]

Where $C_{i}$ is the Compressor input, $D_{o}$ is the decompressor
output and $\epsilon_{auto}$ is a small number (hyper-parameter)
that decreases to zero as the training progresses. 

\section{Text Generation\label{sec:Text-Generation}}

We attempt to achieve the goal of generating text that is indistinguishable
from human generated text using a Generative Adversarial Network \citet{2014arXiv1406.2661G}.
For each level of the DVT (except for the token level), AGENT has
a Generator that generates a text vector, and a Discriminator that
detects whether a text vector was generated via the Generator or the
HE.

\subsection{Generator\label{subsec:Generator}}

To generate text at the paragraph level, we generate a random vector
$V\sim N(\mu_{p},\sigma_{p})^{D_{p}}$ (where p stands for paragraph).
Due to the fact that a randomly generated vector is unlikely to reside
within a region of space that contains coherent text representation
\citet{2015arXiv151106349B}, this vector is passed through a generator
$G_{p}$. $G_{p}$ is composed of L dense layers, where all except
the last one are succeeded by a tanh activation. Following the reasoning
of \citet{2014arXiv1406.2661G}, if $G_{p}$ could be any arbitrary
function, a unique solution would exists where $G_{p}$ recovers the
distribution of paragraph vectors generated by the HE. We aim to select
L in such a way that $G_{p}$ could approximate this solution to a
reasonable degree. The Generators of the other levels of the DVT are
all created in a similar manner.

We also note that there exist an interesting similarity between the
Generator and the Coherence Checker of the same level. While the $G_{level}$
takes a random point and brings it to a region in $\mathbb{R}^{D_{level}}$
where coherent texts ought to reside, $CC_{level}$ takes a point
in $\mathbb{R}^{D_{level}}$ and determines whether or not it is in
that region. For this reason we use the same architecture (L dense
layers) for both. 

\subsection{Discriminator}

\begin{figure*}[t]
\includegraphics[scale=0.3]{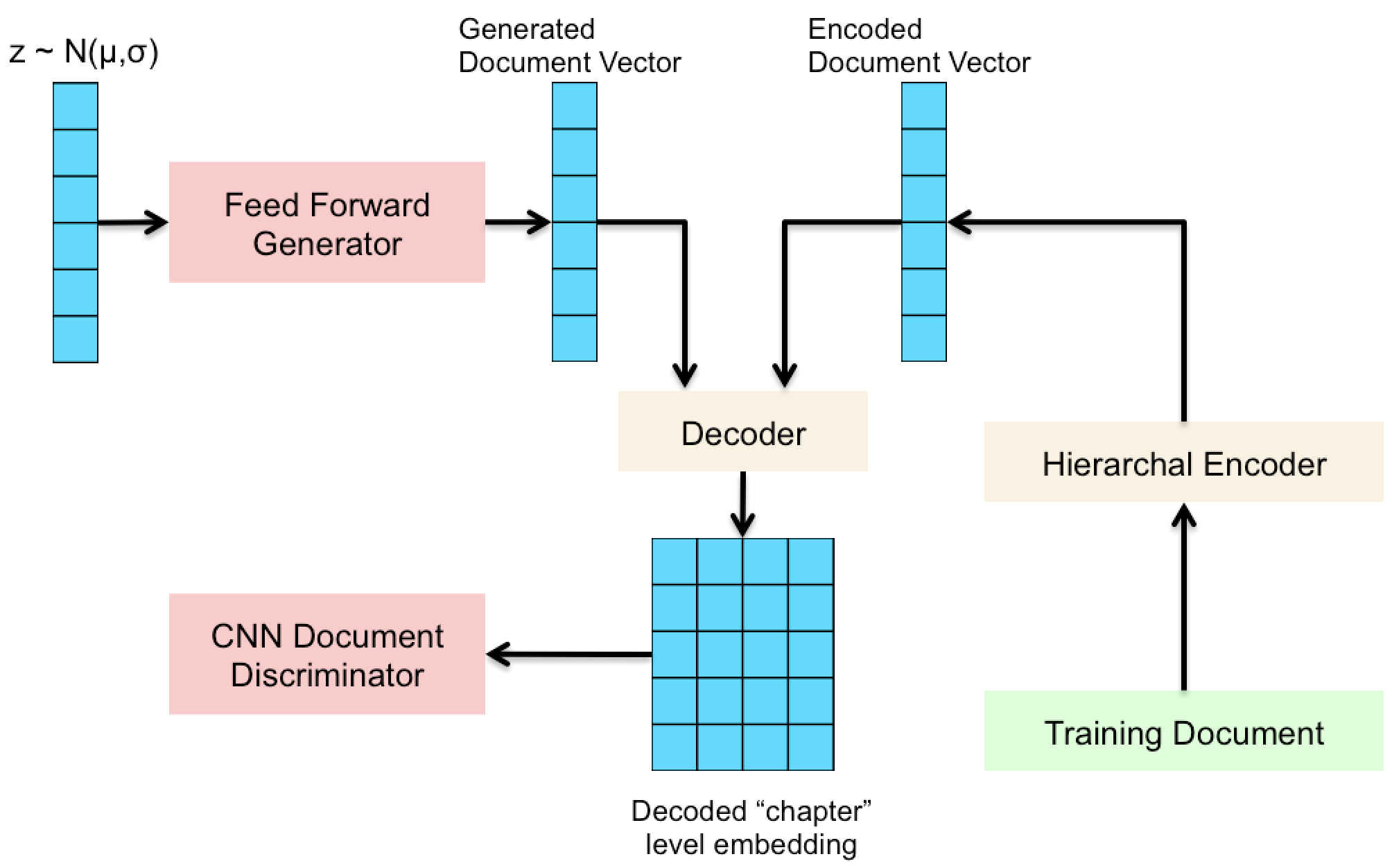}\caption{\label{fig:Generator-discriminator}The Generator-discriminator architecture
of AGENT at the document level}
\end{figure*}

In order to improve the quality of $G_{p}$ we create a Discriminator
denoted as $D_{p}$. As in \citet{2014arXiv1406.2661G}, the Discriminator
aims to classify whether its input was derived from the data or produced
by $G_{p}$, while $G_{p}$ is trained to maximize the loss of $D_{p}$.
However, $D_{p}$ does not directly take the paragraph vector generated
by $G_{p}$ as an input. Instead, this vector is transformed by the
corresponding Decompressor and Decoder to obtain a sentence matrix
for its input. On the other hand, inputs derived from real data are
generated by using the same transformation on paragraphs vectors from
the DVT. The Discriminators' architecture was chosen to be a simple
CNN \citet{D14-1181} as it was proven effective when employed over
LSTM generated word vectors \citep{Zhang2016GeneratingTV}. Though,
unlike \citet{Zhang2016GeneratingTV}, this LSTM is not a part of
the Generator and its weights do not change during the back-propagation
of the Discriminator loss. Instead, its (and the Decoders') weights
are optimized to reconstruct the sentence vectors of the DVT as described
in \ref{subsec:Sequence-Reconstruction}. The Discriminators for the
other levels of the DVT are all created in a similar manner, and figure
\ref{fig:Generator-discriminator} illustrates the above process.

\subsection{Hierarchical Decoding and Copyediting}

After the training of AGENT is done, we use the top level Generator
to create a document vector. This vector is passed to the top level
Decompressor and Decoder, and each resulting vector is passed in turn
to the decoder of its level. By the end of this process, which we
call the Hierarchical Decoding (HD), we are left with a DVT whose
leaves can be translated into tokens by the lowest level decoder.
However, before generating text in that manner, we perform an additional
step, which we refer to as Copyediting. In this step the DVT is repeatedly
traversed level by level, where all nodes of the same level are simultaneously
updated according to the following formula:

\[
N_{i+1}=\epsilon_{edit}\cdot N_{i,MLM}+\left(1-\epsilon_{edit}\right)\cdot N_{i}
\]

where $\epsilon_{edit}$ is a small constant, $N_{i}$ is the current
node after `i' iterations of Copyediting, and $N_{i,MLM}$ is the
same node, replaced by the \textless MASK\textgreater{} vector of
its level and reconstructed using the other (unchanged) nodes of the
same level in accordance with the MLM task (\ref{subsec:Masked-Language-Modeling}).
After updating all the nodes of a level, the lower level of the DVT
is generated from the updated nodes and the traversal progresses.

The motivation behind this step is to use the knowledge we have gathered
during the MLM task of how neighboring texts (sentences in the same
paragraph for example) relate to each other, in order to constrain
the decoded text to exhibit such constraints as well. In other words,
the weights of the MLM component contains knowledge about how subsequent
sentences (in the training examples) behave, and this knowledge might
not overlap completely with the knowledge contained in the HD.

\section{Proper Nouns Data Base\label{sec:Proper-Nouns-Data}}

\begin{figure*}[t]
\includegraphics[scale=0.23]{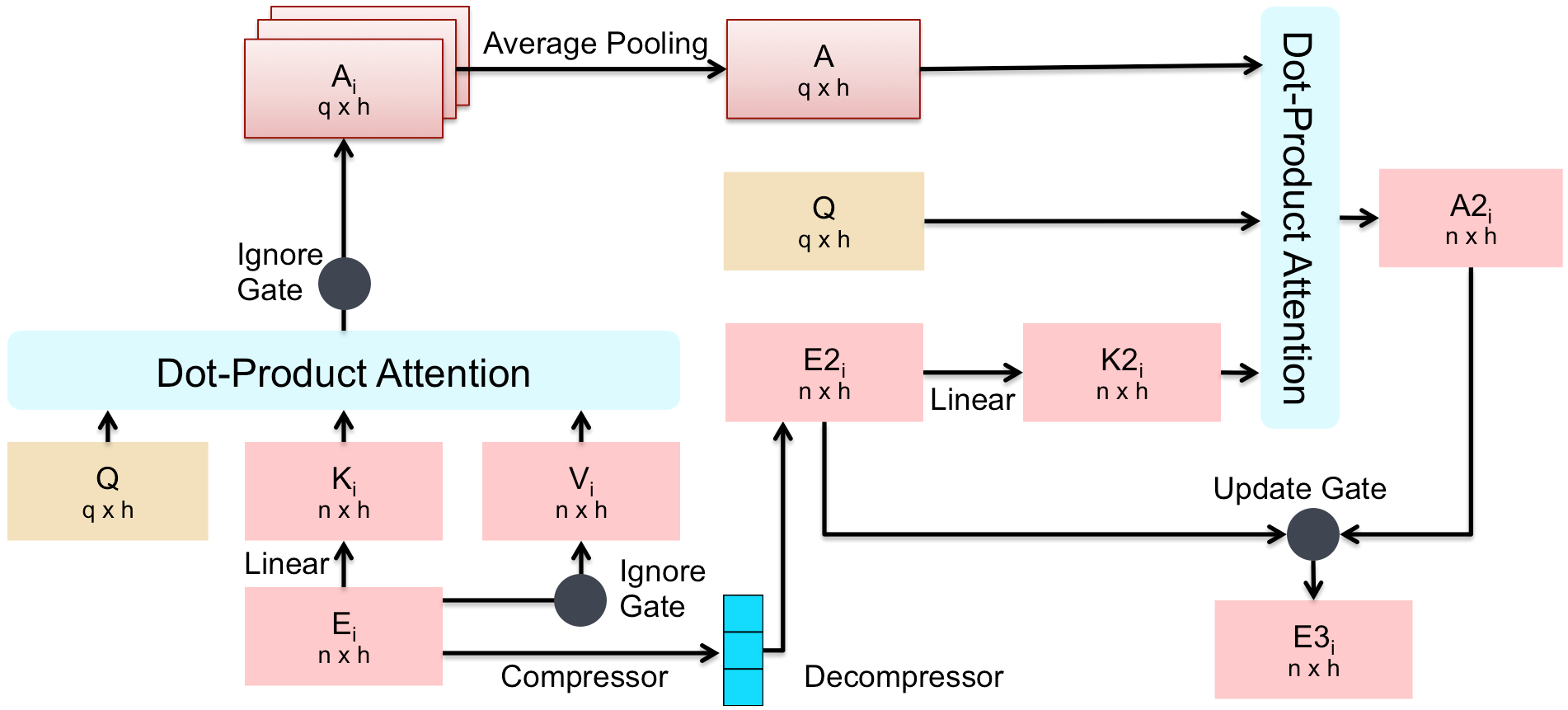}\caption{\label{fig:PNDB}The Proper Nouns Data Base architecture}
\end{figure*}

One challenge in generating a document vector is that the different
chapter vectors that are derived from it need to be consistent. For
example, character names must remain consistent through all chapters
and all of the paragraphs that are derived from these chapters. Though
this challenge can be eased by the using the Coherence loss during
training and by the Copyediting stage during text generation, complete
consistency throughout an entire document might still be very hard
to reach. In addition, many of the sentences in book include duplicate
information such as the name of the main character. While encoding
those sentences into sentence vectors, the HE must make sure that
names are encoded correctly (in such a way that they can be decoded).
This, in turn, requires us to use sentence vector of a dimension that
is large enough to ensure that they can contain all of the names that
might appear in a sentence. 

To address these issues and increase our model's consistency while
reducing the size of its embeddings, we introduce an additional component
to our model that we refer to as Proper Nouns Data Base (PNDB). Figure
\ref{fig:PNDB} illustrates the PNDB and we write to it in the following
manner:
\begin{enumerate}
\item Writes to the PNDB are performed alongside the construction of the
DVT. 
\item We define a question matrix $Q\in\mathbb{R}^{q\times h}$, which is
a learnable model parameter. `q' is a hyper parameter describing the
number of questions our model asks and $h=D_{token}$.
\item For each sentence `i', the HE creates a contextual token embedding
matrix E\textsubscript{i}. For each E\textsubscript{i} we create
a `key' matrix K\textsubscript{i} through a linear projection, and
a `value' matrix V\textsubscript{i} through an Ignore Gate, see section
\ref{subsec:The-Ignore-Gate} for additional details.
\item For each E\textsubscript{i} we create an `answer' matrix A\textsubscript{i}
by applying a dot-product attention in a manner identical to \citet{DBLP:journals/corr/VaswaniSPUJGKP17}
$A_{i}=softmax(\frac{QK_{i}^{T}}{\sqrt{D_{token}}})V_{i}$.
\item A global `answer' matrix A is created by average pooling all the A\textsubscript{i}
matrixes. 
\end{enumerate}
We read from the PNDB the following manner:
\begin{enumerate}
\item Reads from the PNDB are preformed during loss calculations for the
Reconstruction and Masked LM tasks.
\item For each sentence `i' we define E2\textsubscript{i} as the decompressed
token matrix for the Reconstruction task (as illustrated in \ref{fig:PNDB}),
and as the contextual token embedding for the MLM task. 
\item For each E2\textsubscript{i} we create a `key' matrix K2\textsubscript{i}
through a linear projection, and an `answer' matrix A2\textsubscript{i}
by applying a dot-product attention in a manner similar to \citet{DBLP:journals/corr/VaswaniSPUJGKP17}
$A2_{i}=softmax(\frac{K_{i}Q^{T}}{\sqrt{D_{token}}})A$.
\item An embedding matrix E3\textsubscript{i} is created by combining E2\textsubscript{i}
and A2\textsubscript{i} through an Update Gate, see section \ref{subsec:The-Ignore-Gate}
for additional details.
\item The E3\textsubscript{i} matrixes are used as the contextual embedding
matrixes for loss calculations and text generation as described in
\ref{sec:Model-Training} and \ref{sec:Text-Generation}.
\end{enumerate}
While using the PNDB we run the risk of circumventing our model's
learning process due to `label leaking', which in this case means
being able to save a word we are trying to predict to the PNDB, and
then relying on the PNDB to `guess' the missing word without needing
any input from the rest of the model. To avoid such cases we have
made the following decisions for the PNDB architecture:
\begin{enumerate}
\item The Q matrix is a model parameter so we would not be able to `ask'
questions that are specific to a certain document.
\item 'q', the number of possible answers, is set to be much smaller than
the number of tokens in a document.
\item The average pooling layer prevents us from reading answers from the
PNDB that are specific to the i-th sentence (for any i). Thus, to
reduce the overall loss of the model, we are forced to write only
data that has relevance to the entire document to the PNDB.
\end{enumerate}
Alternatively, it is also possible to change `A' to be the average
pooling of all A\textsubscript{i} except for i=j when calculating
loss for the j-th sentence, thus avoiding this issue completely.

\subsection{The Ignore and Update Gates\label{subsec:The-Ignore-Gate}}

While calculating the Masked LM loss for the following example sentence:
``Alice is happy because Bob loves her very much'', we expect that
masking a pronoun (``her''), adverb (``very'') or adjective (``much'')
would make our task relatively easy, while masking a proper noun (``Alice''
or ``Bob'') would make our task harder. However, knowing that Alice
appears in somewhere else in the document can be of great help as
it means she is a character in our story. So, if we are able to save
and retrieve the names of proper nouns in our document, the questions
of ``Who is happy because...'' becomes a multiple choice question
which is presumably easier. Therefore, at the token level, the Ignore
gate was designed to facilitate the saving of proper nouns to the
PNDB.

It has been shown that the attention heads of Transformer-based Encoders
learn to detect part-of-speech (PoS) \citet{2018arXiv180408199S,DBLP:journals/corr/VaswaniSPUJGKP17}.
It follows that Transformer-encoded tokens carry data regarding their
PoS and that this data can be accessed without the use of additional
non linear layers. Therefore, we use a CNN Unigram with `F' filters
to capture this data. These filters are later divided in to groups
of 8 and a softmax activation is applied over each group for each
token. So for each token, each such group `chooses' one of 8 possible
categories. Then, for each token, the results of the `F' post activation
filters are fed in to a single logistic unit. The output of the Ignore
Gate is equal to its input where each token is multiplied by its corresponding
logistic activation. 

The Update Gate is designed to detect tokens in E2\textsubscript{i}
that require additional input from the PNDB in order to have a comprehendable
meaning, and retrieve that input from A\textsubscript{i} when it
is needed. To detect these tokens, we think about the concept of ``require
additional input'' as a PoS, and use the architecture of the Ignore
Gate and its sigmoid unit to detect it. Then, the output of the Update
Gate for th j-th token of E2\textsubscript{i} is a weighted average
of the j-th token of A2\textsubscript{i} and the j-th token of E2\textsubscript{i}
where the weight is determined by the activation of the sigmoid unit.

\subsection{Data Base Generation}

When using the PNDB component during the calculation of the Reconstruction
loss, it becomes necessary to be able to generate values for the `A'
matrix in order to generate a document. To generate the i-th answer
vector in the PNDB, we define G\textsubscript{PNDB}, that performs
the following steps:
\begin{enumerate}
\item The i-th question vector in `Q' is concatenated with the generated
document vector and pass them through a dense layer of size $D_{token}$.
\item The result is concatenated with a random vector $V\sim N(\mu_{token},\sigma_{token})^{D_{token}}$
and the i-th answer vector is obtained using L dense layer, as in
\ref{sec:Text-Generation}.
\item The weights of the G\textsubscript{PNDB} are optimized during the
backpropagation of the Discriminator from \ref{sec:Text-Generation},
which remains unchanged.
\end{enumerate}

\section{Summary \label{sec:Summary}}

In this paper we have presented a new way to encode long texts while
avoiding two of the issues that hinder the performance of classifiers
that are built over SOTA text representation vectors such as \citet{DBLP:journals/corr/abs-1810-04805,Cer:2018aa}.
We reduced the number of operations needed to learn relationships
between different words from $O\left(n^{2}\right)$ to $O\left(n\cdot log\left(n\right)\right)$
due to the hierarchical nature of the DVT, and we have generated text
vectors that can be set to be sufficiently long to represent large
documents. We hypothesize that training AGENT will exhibit the following
steps:
\begin{enumerate}
\item AGENT will learn high quality token representation using the MLM task,
similarly to \citet{DBLP:journals/corr/abs-1810-04805}.
\item Once the token representation is relatively stable, AGENT will attempt
to represent sentences in such a way that:
\begin{enumerate}
\item Tokens can be recovered (to some extent) from their sentence vector
due to the downward (Reconstruction) loss.
\item Coherent sentences reside in a different parts of $R^{D_{sentence}}$
from incoherent sentences due to the upward (Coherence) loss.
\item Sentences with similar meaning (and different wording) should have
similar vectors due to the in-level (MLM) loss. This is because surrounding
sentences can give a lot of information about the meaning of a masked
sentence but very little information about its exact wording and word
order. Therefore, many sentences with the same meaning can be guessed
to be the missing sentence, and in order to reduce the MLM loss their
vectors should be close to each other but far from other, inappropriate
sentences. 
\end{enumerate}
\item Once the sentence representation is relatively stable, AGENT will
attempt repeat step 2 for higher levels of text representation.
\end{enumerate}
The training process of the Generators and Discriminators of AGENT
is characterized by the fact that back-propagating their loss does
not effect the weights of the HE and HD (though a change in either
of them will effect the Discriminators). So one possibility is to
save training time by only training the Generator and Discriminator
pairs after the rest of the network. Another possibility is to only
train the Generators, while the weights of the Discriminators are
set to be equal to their corresponding CC and observe the generated
texts during training.

\bibliographystyle{plainnat}
\bibliography{patrick}

\end{document}